\documentclass[twoside,11pt,theapa]{article}
\usepackage{times}
\usepackage{dsfont}
\usepackage[centertags]{amsmath}
\usepackage{amsfonts}
\usepackage{amssymb}
\usepackage{amsthm}
\usepackage{newlfont}
\usepackage{algorithm,algpseudocode}
\usepackage{rotating}
\usepackage{mathtools}
\usepackage{caption}
\usepackage{subcaption}
\usepackage{graphicx}
\usepackage{authblk}
\title{\textbf{Fast constraint satisfaction problem and learning-based algorithm for solving Minesweeper}}

\author[1]{\textbf{Yash Pratyush Sinha}}
\author[1]{\textbf{Pranshu Malviya}}
\author[1]{\textbf{Rupaj Kumar Nayak}}
\affil[1]{International Institute of Information Technology Bhubaneswar, India}
\date{}

\begin{document}


\maketitle

\begin{abstract}

Minesweeper is a popular spatial-based decision-making game that works with incomplete information. As an exemplary NP-complete problem, it is a major area of research employing various artificial intelligence paradigms. The present work models this game as Constraint Satisfaction Problem (CSP) and Markov Decision Process (MDP). We propose a new method named as dependents from the independent set using deterministic solution search (DSScsp) for the faster enumeration of all solutions of a CSP based Minesweeper game and improve the results by introducing heuristics. Using MDP, we implement machine learning methods on these heuristics. We train the classification model on sparse data with results from CSP formulation. We also propose a new rewarding method for applying a modified deep Q-learning for better accuracy and versatile learning in the Minesweeper game. The overall results have been analyzed for different kinds of Minesweeper games and their accuracies have been recorded. Results from these experiments show that the proposed method of MDP based classification model and deep Q-learning overall is the best methods in terms of accuracy for games with given mine densities.

\end{abstract}

\section{Introduction}\label{Introduction}
Minesweeper is a single-player game where the player has been given
a grid minefield of size $p \times q$ containing $n$ mines where
each block of the grid contains at most one mine. These mines are
distributed randomly across the entire minefield and their locations
are not known to the player. The goal of the game is to uncover all
blocks which do not contain a mine (i.e., all safe blocks). If a
block containing a mine is uncovered, the player loses the game.
Whenever a safe block is uncovered, it represents the number of
mines in its $8$-neighbors containing mines. In the human version of
the game, when this number is zero the game automatically uncovers
all of the 8-neighbours of such a block. Generally, the first block
uncovered by the player does not contain a mine. Analyzing the given
pattern and arrangement of these boundary digits, the player has to
figure out the position of the safest block that can be uncovered.
The player can also mark a block with a flag if he/she decides that
the block is mined. In this way, the number of mines left to be
discovered is updated and the player can continue to play until all the
free blocks are revealed. In some situation, if there exists one
than one block with similar chances of containing a mine, but not
with the surety of absence of mine, the player has to randomly
choose the next block to uncover. As a result of this ambiguity, the
player can lose the game even if he applies the most optimal
strategy. As it is computationally difficult to make an optimal
decision without considering all options, Minesweeper is an
NP-complete problem \cite{kaye} which is an intriguing area of
research. Many researchers have already described the formulation
techniques that can be employed for this game and improved upon the
previous ones as discussed in the next section.


\section{Related Work}\label{Related Work}
A strategy called the single-point strategy considers only one instance
of the uncovered block and finds the safest move from its immediate
neighbours. Its implementation model may vary but the single point
computations are common practice in them. As a result, this strategy
struggle with larger dimensional games or games with higher mine
densities. Therefore, other strategies can be used to reconcile the
limitations of single-point techniques. CSP is another prominent way
of formulating the Minesweeper game in a mathematical form and
several algorithms \cite{aima} exist to solve these CSPs. CSPs are
the subject of intense research in both AI and operations research
since not only it overcomes the problems in single-point strategy,
but the regularity in their formulation provides a common basis to
analyze and solve problems of many unrelated families
\cite{alldif,dfs}. MDP is another way of formulating games into
state-action-reward form. This way has been extensively practised
nowadays to formulate games \cite{atari}. We also use such
a formulation and combines it with the CSP formulation to achieve even
better results.




Realizing the Minesweeper game as a CSP as described in \cite{stud}
is one of the basic ways that is employed as discussed. The thesis
by David Becerra \cite{becerra} tests and talks about several
different approaches are taken to solve the Minesweeper game and comes
to the conclusion that CSP methods work as efficiently as any other
formulations of the Minesweeper game.

Our method for solving CSPs is in line with the look-ahead method
proposed by Dechter and Frost \cite{back} that avoids unnecessary
traversal of the search space to result in a more optimized method
for dealing with CSPs.


The work of Bayer et al. \cite{ken}
describes another approach that uses generalized arc consistency and
higher consistencies (1-RC \& 2-RC) to solve the CSP generated by
the Minesweeper to assist a person playing a game of Minesweeper.
Both of these approaches are inefficient when constraints are not
very scalable for larger boards.

Nakov and Wei \cite{nw} describes the Minesweeper game as a
sequential decision-making problem called Partially Observable
Markov Decision Problem (POMDP) and convert the game to an MDP,
while also reducing the state space.

Approaches have been carried out like using belief networks by Bonet
and Geffner \cite{jair} that consider a set of partially observable
variables and considers the possible states as beliefs.

In the research by Castillo and Wrobel \cite{multir} authors
described methods to learn the playing strategy for Minesweeper not
just by inferring from a given state but also by making an informed
guess that minimizes the risk of losing. They used Induction of
Logic Program (ILP) techniques such as macros, greedy search with
macros, and active inductive learning for the same. This is very
different from another machine learning approach that we have used
which is also very promising.

Sebag and Teytaud \cite{uct} have attempted to estimate the belief
states in the Minesweeper POMDP by simplifying the problem to a
myopic optimization problem using Upper Confidence Bounds for
Monte-Carlo Trees (UCB for Trees). While this method is reasonably
accurate for small boards, it is too slow for larger board sizes
making it inadequate. As described in research of Legendre et al.
\cite{hcsp}, authors have used heuristics with computing
probabilities from CSP called HCSP to study the impact of the first
move and to solve the problem of selecting the next block to uncover
for complex Minesweeper grids based on various strategies. By
preferring the blocks which are closest to the frontier in case of a
tie on the probability of mines, they got the better results for
various sizes of Minesweeper matrices. We have also used a variant
of HCSP by using Manhattan distance for measuring closeness to the
boundary.

In the research given by Buffet et al. \cite{ucthcsp}, authors
combined the two methods i.e., UCB for trees and HCSP as described in
\cite{uct} and \cite{hcsp} respectively to improve the performance
of UCB to be used not only for small boards but also for big
Minesweeper boards. This improves performance by a lot.

Recent works by Couetoux et al. \cite{conbelief} have tried to solve
Minesweeper by directly treating it as a Partially Observable MDP
(POMDP).

However, as pointed out by Legendre et al., Minesweeper is actually
a Mixed Observability MDP (MOMDP) and several improvements can be
made to the current solvers to increase performance.

In the research of Gardea et al. \cite{gardea} the authors have
taken an approach to utilize different machine learning and
artificial intelligence techniques like linear and logistic
regression and reinforcement learning.

Q-learning \cite{watkins1992q} is a reinforcement-learning algorithm
which has been used extensively in games \cite{atari,mnih2015human}
to solve MDPs. We have also used a modified version of the Q-learning
for solving the above MOMDP efficiently.



In this paper, we have described our methods of creating an
automatic solver for Minesweeper using several techniques by
gradually improving upon its previous ones. We have also used a
modified version of Q-learning for solving the above MOMDP efficiently.
In section 2, we describe how the game is formulated into CSP and
then in an MDP. We describe techniques to play the game using both
the CSP \& MDP formulation. Also, we have described three ways to
obtain the probabilities of each block containing a mine as well as
a method to find blocks about which we can deterministically say if
they contain a mine. After that, we give several methods, both
hand-crafted and machine learning-based to choose blocks in the case
that there is no deterministic solution. In section 4 we described
the implementation of these methods and present the results
graphically and a comparison table with existing popular methods. We
conclude the paper in section 5.

\section{Formulations}\label{Formulation}
Prior to the detailed explanation of algorithms and machine learning
techniques that we used, in this section, we are going to describe
how the minesweeper game is first realized as a playable
mathematical form. These forms and notations then allowed us to
design and improve upon the existing algorithms, particularly for
this game for better results. This paper particularly uses two
different formats, expressing the game as a CSP or as an MDP.
\subsection{CSP formulation}

CSP \cite{aima} is a natural and easy way for the formulation of
minesweeper, as it accurately captures the intuition of finding
which blocks may or may not contain a mine. We use the fact that
each uncovered block shows the number in its 8-neighbour region in
order to build a CSP from the minesweeper game. These numbers put a
constraint on the covered blocks in the neighbourhood that how many of them should have a mine. (We also know the total number of mine
present in the game. This information will be used as heuristics
discussed later in the paper.) If we treat the covered blocks as
boolean variables (with $0$ representing a safe block and a $1$
representing a mine) we can use the uncovered numbers as constraints
on such variables. For example, a two uncovered block will constrain
the sum of all of the variables in its eight neighbours to two.
Assuming the uncovered number of every block at $(i,j)$ represented
by num$_{i,j}$ and the boolean variables represented by
mine$_{i,j}$, the constraints can be written as (adaptively for
blocks on the border):
\begin{equation}
\label{cspeq}
num_{i,j} = \sum_{k=i-1}^{i+1} \sum_{l=j-1}^{j+1} mine_{k,l}~~~if(k\neq i~\&~l\neq j)
\end{equation}

For several $(i,j)$ we will obtain a set of the equation in the form
of equation (\ref{cspeq}). We can re-write this set of equations in
a matrix format as follows:
\begin{equation}
\label{cspeqn}
AM=N,
\end{equation}
Where $M$ is the vector denoting a list of uncovered variables, $N$ is the
list of all applicable num$_{i,j}$ and $A$ is a binary coefficient matrix
obtained from equation (\ref{cspeq}).

\subsection{MDP formulation}
Markov Decision Process (MDP) is formulating the minesweeper game as $5-tuple$ of ${(S,A,T,R_{\cdot }(\cdot ,\cdot ),\gamma )}$ where
\begin{itemize}

\item $S$ is a set of all the possible states of the game that represents all the possible board configurations at any time.
It includes a special state $s_{init}$ which is the initial state of
the game as well as the final states $s_{lose}$ and $s_{win}$ which
represents the state when the game is lost and the state of winning
the game respectively.

\item $A$ is the set of all actions that can be performed at any time on a
state $s$ to go to some next state $s'$. For minesweeper, this
represents all the covered blocks that can be uncovered to go from the
current board state to the board state after the block has been
uncovered.
\item $T$ is the set of transition triplets $(s,a,s')$ which is the probability
of going to the next state $s'$ from the current state $s$ by doing
the action $a$. In our consideration, the probability for $a$ being $1$ means the player is surely going to lose.
\item $R$ is the reward function $R(s,a)$ representing the reward to
be received
  after performing action $a$ on a state $s$.
Our model rewards an action $a$ according to the number of blocks
that are deterministically uncovered after it is performed, as we want to
encourage the model to choose the action which would uncover the maximum number of blocks with the least chance of losing the game. 

\item $\gamma$ is the discount factor according to which the future
rewards are reduced.
\end{itemize}

A POMDP is a more general way of formulating the game as it
captures the intuition that the entire state is not observable.
POMDPs are solved by using belief states, which are generally
computed using simulations or other tree search methods with
$b_{s,a}=b(s,a,s')$ where $b_{s,a}$ is the belief on state $s$ on
taking action $a$ and observing state $s'$. This entire method,
however, overlooks the fact that a part of the state is fully
observable making the problem a Mixed Observability MDPs (MOMDPs).
MOMDPs are generally easier and faster to simplify and solve than
POMDPs. Our method uses this fact while ignoring the non-observable
part of this MOMDP in the belief state to further simplify this
problem. Since we neglect both the hidden part of the state and the
future observation in this belief, we call it a sub-state denoted
by $s_a=b(s, a)$. Here $b$ is a function that
takes a section of the board as a subset, based on the action.
This sub-state is dependent on the current state
and the action to be performed only. The details of how and what
sub-states have been used are given in section 3.2.

\section{Solving Method}\label{Solving}
This section contains strategies to find the next best block to
uncover. Our solution is based on a two-step process. The first step
is called Deterministic Solution Search (DSS), which evaluates
solutions for the backbone variables (whose value is the same
in all solutions) if they exist for equation (\ref{cspeqn}).
DSS collects all deterministic variables which satisfy the
constraints in the equation (\ref{cspeqn}) and assigns them to
be uncovered. The other non-deterministic variables are ignored.

In this way, we do not have to go for any other methods to find the
safest block each time. This method proves to be very effective as
it has a worst-case time complexity to $\mathcal{O}(n^2m)$. The
steps to find deterministic variables are given in Algorithm
\ref{DSS}.

\begin{algorithm} \caption{: Reduce(A,N,$j$,value)}
  \label{Reduce}
  \begin{algorithmic}[1]
    \If{$value$ is $0$}
      \State Set column $j$ of $A$ to $0$
    \ElsIf{$value$ is $1$}
      \State Reduce value of $N_j$ by $1$
      \State Set column $j$ of $A$ to $0$
    \EndIf
  \end{algorithmic}
\end{algorithm}
\begin{algorithm} \caption{: DSS(A,M,N)}
  \label{DSS}
  \begin{algorithmic}[1]
    \State $Detrmined \leftarrow $ \textit{empty list}
    \State $loops \leftarrow $ Any significanly large number, say $10$
    \For{$loops$ number of times}
      \For{Each row $i$ in $A$}
        \State $Variables \leftarrow$ \textit{Number of variables in $A_i$}
        \If{$Variables$ is $0$}
          \State Delete row $A_i$ from $A$
        \ElsIf{$N_i$ is $0$}
          \For{every variable $var_{i,j}=1$ in $A_i$}
            \State $Reduce(A,N,j,0)$
            \State Add variable $M_{j}$ with value $0$ to to $Determined$
          \EndFor
        \ElsIf{$Variables = N_i$}
          \For{every variable $var_{i,j}=1$ in $A_i$}
            \State $Reduce(A,N,j,1)$
            \State Add variable $M_{j}$ with value $1$ to to $Determined$
          \EndFor
        \EndIf
      \EndFor
    \EndFor
    \State Return $Determined$
  \end{algorithmic}
\end{algorithm}

Now, if Algorithm \ref{DSS} returns a non-empty vector there exists
some determined blocks which can directly uncover them or flag.

But if it is found that there is no such deterministic variable, we
move to the next step. The primary approach of a player is to select
the next move based on current knowledge of the state and thus we
estimate the probabilities of each block being safe or unsafe.

\subsection{From CSP}\label{prob}
We propose approaches to find a set of feasible
solutions for the CSP formulated using equation (\ref{cspeqn}) using which the probabilities
for non-deterministic variables are estimated.
We generate a solution set by either using a simple backtracking method or by our
proposed DSScsp algorithm where both algorithms require traversing of a
tree.
\begin{itemize}
\item The first one is to use backtracking that constructs a pruned
tree where the path for every deepest leaf node is one possible
solution. The tree is traversed recursively to gather solutions that
satisfy the given set of constraints or equation (\ref{cspeqn}). This was implemented mainly to compare with our proposed method.
\begin{algorithm} \caption{: DSSCSP(A,M,N)}
  \label{DSScsp}
  \begin{algorithmic}[1]
    \State Static $S \leftarrow$ empty list
    \If{$A$ is a Zero-Matrix}
      \State $M$ is a possible solution, add it to $S$
    \EndIf
    \State $i = argmax_{i \in M}(\sum_j^{M_i}j)$; where $M_i$ is $i^{th}$ column in $M$
    \State $A',M',N'$ = $A,M,N$
    \State $M'_i$ = $0$
    \If{$A'M'=N'$ is possible}
	    \State $Reduce(A',N',i,M'_i)$
	    \State Run $DSS(A',M',N')$ for further reduction
	    \State $DSSCSP(A',M',N')$
	  \EndIf
    \State $A'',M'',N''$ = $A,M,N$.
    \State $M''_i$ = $1$
    \State \If{$A''M''=N''$ is possible}
	    \State $Reduce(A'',N'',i,M''_i)$
	    \State Run $DSS(A'',M'',N'')$ for further reduction
	    \State $DSSCSP(A''',M'',N'')$
	  \EndIf
    \State Return $S$
  \end{algorithmic}
\end{algorithm}

  \item
DSScsp, as given in Algorithm \ref{DSScsp}, uses DSS, in the
Algorithm \ref{DSS}, at each step, to quickly reduce the backtracking
search space.
The depth of the tree is drastically reduced as several
variables are resolved in a single node due to DSS. This leads
DSScsp being randomized and much faster than simple backtracking in general. The
$Reduce$ function in Algorithm \ref{Reduce} is used in both
Algorithm \ref{DSS} and Algorithm \ref{DSScsp}.
\end{itemize}

Using one of these ways, we now have a set of solutions, or a
solution matrix $S$ that satisfy the given constraint equation
(\ref{cspeqn}). To limit the time taken to find these
equations, we have employed a strategy of just computing a random
large enough subset of all the possible solutions over the full set
of solutions. We do this by setting a maximum number of solutions, a
maximum depth as well as a maximum number of iterations. The
guarantee of the sample being random comes from the fact that, in
DSScsp we assign values $1$ or $0$ randomly.

This set of solutions allows us to compute the probabilities of a
block containing a mine.
\begin{equation}\label{probs}
P^\top=
      \begin{bmatrix}
        \displaystyle\frac{\sum_{i=1}^{s} S_{1,i}}{s}, ~~\displaystyle\frac{\sum_{i=1}^{s} S_{2,i}}{s},~~\dots~~ ,\displaystyle\frac{\sum_{i=1}^{s} S_{m,i}}{s} \\
      \end{bmatrix}
\end{equation}

where, $S$ represents the binary solution matrix and $s$ is the
number solutions generated. In $S$ matrix, each row corresponds to a
unique solution to equation (\ref{cspeqn}) i.e., $S_{i,j}=1$ only if
block $j$ contained a mine in solution $i$.

The simplest way to proceed with the game after obtaining probabilities
is to uncover the block with the least probability
of having a mine. But there are ways to improve. We developed several
heuristics in addition to these probabilities that play a crucial
role and employed machine learning algorithms to have faster execution with more accuracy.

However, in Minesweeper, even the supposedly best move
picked by any heuristic, might result in losing the game.

\subsection{Heuristic approaches}\label{heu}


We have built heuristics that consider the problem using several
other parameters and formulations like the size of the board, total
number of mines, flags used, the location of blocks etc. which may
affect the results. We have used machine learning to find the best
heuristics with these parameters.


\subsubsection{ Manhattan Distance}\label{manD}
Legendre et al.\cite{hcsp} consider a heuristic that the blocks
closer to the edges of the board are likely to be more informative
about blocks involved in equation (\ref{cspeqn}) as compared to the
farther ones. We consider the same and use the manhattan distance
between block at $(i,j)$ to the nearest edge of the board. We also
relax the rule of choosing the block with minimum probability.
Instead, we choose a list of blocks $B_{s}$ such that $\forall m
\in B_{s} : P_{s} \leq min(P) + 0.05 $ where $min(P)$ is the
minimum probability in $P$, and $P$ is the vector of probabilities 
of a block containing a mine (\ref{probs}). Now that we have a reduced list of safe
blocks, we choose the one with the minimum manhattan distance from
the minefield boundary. We uncover this block now.

\subsubsection{ Supervised Learning}\label{ml}

This method follows a machine learning approach in which we define
the given situation as a $state$ and uncovering a block as a
$action$. It is a binary classification model that is trained on the
different states and their corresponding actions taken in
Minesweeper game. We collect the data from the games played randomly
by the previous versions. The aim is to classify the $action$ as
safe or unsafe for the current $state$. We divide the collected data
as $x_{train}$ and $y_{train}$ as follows.
  \begin{itemize}
\item $x_{train}$: The input part $x$ that describes the information we 
know from the current $state$ and an $action$. It consists of the following:
\begin{itemize}
\item Size of the board ($p,q$)
\item Number of mines
\item Coordinates of the covered blocks ($i,j$): These are the coordinates
of the covered blocks in the neighbor (variable boundary) of
uncovered blocks. We also consider $3$ random covered blocks in
current $state$.
\item Probability ($p$): The probability is the same as obtained from DSScsp algorithm
if the covered block lies in the variable boundary in the current
$state$. Otherwise, its value is assigned to $0.5$ as there is no
information given for that block.
 \item Size of $M$.
\item Variable with minimum probability ($index(p_{min})$):
It is the index ($action$) of the covered block from the variable boundary
with minimum probability of having mine.
\item Score(corner/edge/middle): It is the additional heuristic
we use in this method. We take the probability of a block to have no
neighboring mine into account so that more blocks can uncover and we
get better and more detailed view of the game from the new $state$.
In other words, we are considering the next immediate $state$ of the
minesweeper game just after we click a block. This score varies as
per the orientation of the new block on the board: \\If $f$ is the
number of flags already used and $l$ is the number of uncovered blocks left,
\begin{equation}\label{score}
\begin{split}
Score(M_{ij}=0~|~(i,j) & = corner) = 1 - (\frac{m-f}{l})^{4} \\
Score(M_{ij}=0~|~(i,j) & = edge) = 1 - (\frac{m-f}{l})^6 \\
Score(M_{ij}=0~|~(i,j) & = middle) = 1 - (\frac{m-f}{l})^8
\end{split}
\end{equation}
As corner block will have the maximum score, it will uncover the
larger area of the board that eventually increase chances of winning
the game.
\end{itemize}
\item $y_{train}$: The output part consists of a single column corresponding to
each data point in $x_{train}$ as $y$. Its value is $1$ if the
block turns out to be safe or else it is $0$.
\end{itemize}
We then propose to train two machine learning models with the above
data.


\textit{XGBoost: Single-pass Classification}
Here, we have implemented binary classification of above
state-action pair represented as $x_{train}$ into classes of win
and loss as $y_{train}$. We use single-pass eXtreme Gradient Boost
(XGBoost) \cite{xgb} classifier to do this. It is an ensemble
technique that is used to build a predictive tree-based model for
handling sparse data where the model is trained in an additive
manner. Here, new models are created that predict the residuals or
errors of prior models and then added together to make the final
prediction. So, instead of training one strong learning algorithm
XGBoost trains several weak ones in a sequence until there is no
further improvement. The objective of XGBoost is based on loss
function $L(\theta)$ (we use logistic for binary classification) for
prediction and a regularization part $R(\theta)$ (we use L2 regularization) that depends on
the number of leaves and their prediction score in the model that
control its complexity and avoids overfitting. Let the prediction
$y^{pred}_{i}$ for a data-point $x_{i}$ be,
      \begin{equation}\label{yi}
      y^{pred}_{i}~=~ \sum_{j}^{} \theta_{j} x_{ij}.
      \end{equation}
      Here $\theta_j$\ are learned by the model from data. Considering 
      mean squared error, loss function for actual value $y_{i}$ will be,
      \begin{equation}\label{loss}
      L(\theta)~=~ \sum_{i}^{} (y_{i}~-~y^{pred}_{i})^2.
      \end{equation}
      Hence, the objective function for XGBoost mathematically is,
      \begin{equation}\label{xgboost}
      Obj(\theta)~=~ L(\theta)~+~R(\theta).
      \end{equation}
We optimize the above objective function by training from the given
data and the model tune its parameter accordingly.
When the model is trained with the data, at every new $state$ we
test the above data as input to and the model returns a predicted
score, $y^{pred}$, such that $0\leq y^{pred}\leq1$, where, larger
the value of $y^{pred}$, more is the chance of that block to be
safe. Hence for each $state$, we select the block with the maximum
score, to uncover and proceed to another state.

\textit{Neural Network:} Here, classification
is done by iteratively generating a small batch of data and training
the model with it. This method allows the model to learn and adapt
according to the randomly generated data. The model is a simple
multi-layer neural network for binary classification that takes $x$
as its input and returns $y$ as output. Table \ref{bnn} describes
the neural network layers, the number of neurons in them and the
corresponding activation functions.

We used the Relu activation function for all layers except the output layer
that uses the Sigmoid activation function as $y$ is between $0$ and $1$.
Here, the loss was evaluated using binary cross-entropy for its
logistic behaviour and optimizer as Adam. These configurations were
done after experimental validation. Our model of iterative
classification was inspired by the experience-replay training
methods. In this, however, we just train the model iteratively in
a given number of episodes by generating the training data using the
previously made model. Algorithm \ref{v5.5} describes the iterative
process of training the neural network.
\begin{table}[h!]
      \centering
      \caption{Neural Network} \label{bnn}
      \begin{tabular}{|c|c|c|}
       \hline
      \multicolumn{ 1}{|c|}{{\bf Layer}} & {\bf Number of Neurons} & \multicolumn{ 1}{|c|}{{\bf Activation Function}} \\
      \hline
         Input &   sizeof(x) & ReLU\\
      \hline
         Hidden1 &  sizeof(x) & ReLU\\
      \hline
         Hidden2 &  5 & ReLU\\
      \hline
         Hidden3 &  5 & ReLU\\
      \hline
         Output &  1 & sigmoid\\
      \hline
      \end{tabular}
      \end{table}
      \begin{algorithm}\caption{: Iterative Classification}\label{v5.5}
        \begin{algorithmic}[1]
        \State Import $weights$ from current $model$
        \State $n_{episodes} \leftarrow$ number of episodes
        \State $batch \leftarrow 10$
        \For{each $episode$ from $1$ to $n_{episodes}$}
          \State $x_{train},y_{train} \leftarrow make\_data(batch)$
          \State Train model with $x_{train}$ and $y_{train}$
          \State Save updated $model$ and $weights$
        \EndFor
        \end{algorithmic}
      \end{algorithm}

\subsubsection{Q-Learning}\label{Q-rl}
We use the MDP formulation as described in section 2.2 as a
heuristic. As stated there, we have avoided state space explosion by
using sub-states instead of belief states. The advantage of using
a sub-state over a belief state is that a sub-state can be calculated only
using the current state and the expected action. Belief states, on
the other hand, need to be calculated by trying to predict the
observation, often requiring expensive tree search methods to be
computed. It should be pointed out that while our actual formulation
is a MOMDP, we have ignored the unknown belief part to improve on
computability. In our formulation, these sub-states are of a fixed
size $sub \times sub$ with the centre of the sub-state representing
the block to be uncovered (i.e., the action). We then use deep
Q-learning to solve the MDP formulated earlier by learning a
Q-function to predict expected discounted reward and we then choose
the action with the maximum expected discounted reward.

In Q-learning, we train a model to learn a Q-function given its
inputs. The form of Q-function used in our application is given in
the equation (\ref{Ql-eq}). The Q-function will take the same amount
of time for every board configuration irrespective of the size of
the board. This makes it such that the limitation here is the number
of possible actions which is the number of uncovered blocks. This
Q-learning can find the next action in $\mathcal{O}(pq)$ where $p$
and $q$ are the dimensions of the given board.
computation. The transition rule of Q-learning (SARSA) is given as:
\begin{equation}\label{Ql-eq}
 Q(s_{a},a)=R(s_{a},a) + \gamma \max_{a'} Q(s'_{a'},a')
\end{equation}
Here, $s_{a}$ represents the sub-state of state $s$ for choosing any
action $a$, $s'$ and $a'$ represents the state at after choosing
action $a$.

However, the problem exists that the direct sub-state is not very
representative of the expected reward for any action. For this, we
use a score as defined in equation (\ref{P-sc}) to represent every
covered block in the sub-state. The score is designed such that it
retains information of its safe probability as well as its
location. For a number to represent location we have used the
formula in equation (\ref{score}).
\begin{equation}\label{P-sc}
Sc_{i,j} = \alpha \times P_{i,j} + (1 - \alpha) \times score_{i,j}
\end{equation}
Here P$_{i,j}$ represent the probability of mine being present at
$(i,j)$, score$_{i,j}$ represent the location score and $\alpha$ is
a variable to be chosen such that the Sc$_{i,j}$ represents the best
possible score.
We have used any invalid values to represent an uncovered block. In
order to decrease bias in $\alpha$, we have defined it as a linear
function of board dimension ($p$ \& $q$) and mine ratio $\frac{n}{p
\times q}$.
\begin{equation}\label{linreg}
\alpha = \theta_1 \times p + \theta_2 \times q+ \theta_3 \times
\frac{n}{p \times q} + \theta_4.
\end{equation}

To obtain an $\alpha$ which would give us a good representation of
the score, we obtained win ratios of different values of $\alpha$ by
playing with the heuristic given in section \ref{manD} on the vector
of $Sc$ instead of $P$.   We then performed linear regression with
the $\alpha$ which maximized win ratio as the target to obtain the
values of $\theta_{1}$ to $\theta_{4}$. By using the method
described above, we can obtain scores $Sc_{i,j}$ for every block
$(i,j)$. Using these scores, we can find a sub-state of size $sub
\times sub$ for each action $a$ where each element of the sub-state
is the score in $Sc$. We then defined the immediate reward
$R(s_a,a)$ for any action $a$ as in equation (\ref{reward}).
\begin{equation}\label{reward}
R(s_a,a)=\frac{\textrm{number of newly opened blocks because of
action~}a}{(p \times q)-n}.
\end{equation}

We then ran several simulations to find the net discounted expected
reward $Q(s_a,a)$ for several action $a$ sub-state $s_a$ pairs. A
neural network of the configuration given in Table \ref{rl} was then
trained to predict the Q-function. As done in section \ref{ml},
both single-pass training and iterative training methods were used.
We then choose the action which maximizes the expected discounted
reward $Q(s_a, a)$.
\begin{table}[h!]
      \centering
      \caption{Network for Q-function} \label{rl}
      \begin{tabular}{|c|c|c|}
       \hline
      \multicolumn{ 1}{|c|}{{\bf Layer}} & {\bf Number of Neurons} & \multicolumn{ 1}{|c|}{{\bf Activation Function}} \\
      \hline
         Input &   $sub \times sub$ & ReLU\\
      \hline
         Hidden1 &  $sub \times sub$ & tanh\\
      \hline
         Hidden2 &  $sub \times sub$ & linear\\
      \hline
         Hidden3 &  $sub \times sub$ & linear\\
      \hline
         Output &  $1$ & tanh\\
      \hline
      \end{tabular}
\end{table}

\section{Experiment}\label{Experiment}

For the experiment, we implemented the minesweeper game from scratch
along with a playable interface in C++. We then built the various
versions of the solver in which every version (except version $6.5$
and $3.0$) is an improvement over the previous version with a new algorithm
or tuned parameters. The versions are numbered as given in Table
\ref{versions}.
\begin{table}[h!]
      \centering
      \caption{Versions with algorithm description} \label{versions}
      \begin{tabular}{|c|c|}
       \hline
      \multicolumn{ 1}{|c|}{{\bf Version Number}} & {\bf Algorithm} \\
      \hline
         $1.0$ &   Backtracking \\
      \hline
         $2.0$ &  $1.0$+DSS \\
      \hline
         $2.5$ &  $1.0$ limited + DSS \\
      \hline
         $3.0$ &  DSScsp + DSS \\
      \hline
         $3.5$ &  DSScsp limited + DSS \\
      \hline
         $4.0$ &  $3.0$ + Manhattan Distance \\
      \hline
         $4.5$ &  $3.5$ + Manhattan Distance \\
      \hline
         $5.0$ &  $4.5$ + Supervised Learning using Single-pass Classification \\
      \hline
         $5.5$ &  $4.5$ + Supervised Learning using Iterative Approach\\
      \hline
         $6.0$ &  $4.5$ + Q-Learning using Iterative Classification\\
      \hline
         $6.5$ &  $4.5$ + Q-Learning using Single-pass Classification \\
      \hline
      \end{tabular}
\end{table}

By creating several different versions of the solver we were able to
see the performance improvement caused by each subsequent algorithm.

\subsection{Formulating as CSP}

All versions from $2.0$ to $4.5$ were implemented in C++ for faster
performance. Backtracking with DSScsp was implemented iteratively to
avoid stack overflow as version $2.0$. Version $2.5$ is the limited
traversal of the search tree by limiting the depth and width to be
traversed in the search tree. The width is limited by limiting the
number of solutions to $100$. The depth is limited by limiting the
number of swaps, or the number of edges starting from the root node
of the tree to $1000$. We choose the number of solutions as $100$ as
it gives reasonably accurate results, in much less time.

Version $3.0$ is the implementation of DSScsp Algorithm \ref{DSScsp}
along with Algorithm \ref{DSS} to find the solution set. This method
finds the same solution set as simple backtracking. Hence, it has
the same accuracy as that in Version $2.0$. But, it is much faster
in terms of time taken to solve the Minesweeper game.

Version $3.5$ is the limited traversal of the search tree traversed
in Version $3.0$. The time taken is reduced by limiting the depth
and breadth of the search tree. The limitations are chosen such that
accuracy and amount of time are similar in the case of version $3.5$ and
version $2.5$. We have set the limit to the number of solutions to
$100$ and limit to the depth of the tree to 300, such that the
accuracy of Version $3.5$ remains almost similar to version $3.0$
while reducing the total amount of time.

Versions $4.0$ and $4.5$ are the extensions of versions $3.0$ and $3.5$
respectively. They include the heuristic rules defined in section \ref{manD}.

\subsection{Formulating as MDP}
We used the machine learning method to perform classification that
utilizes the information like $state~-~action$ pair and their
corresponding $result$. We denote this information as one
data-point. The data was collected by playing games using version
$4.0$, with x-dimension of board i.e., $p$, varying from $5$ to $30$
and y-dimension i.e., $q$, as $0.5 \times p$ to $p$. For each board,
we generated at most $100$ different mine distribution possible in
it. The mine ratios for these boards is varied from $5\%$ to $30\%$.

\subsubsection{Classification}

For version $5.0$, we collected this ample amount of data of around
$23.3$ million and trained it using XGBoost classifier. We used the
xgboost library in Python to implement this model and save the
trained model as a binary file. This binary file is then accessed in
C++ to play the game with version $5.0$.

For version $5.5$, we collected a batch of data of more than
$0.7$ million samples similar to the previous version and
trained the binary neural network (Table \ref{bnn}). We
saved the updated neural network model and then again generated a
new batch of data to repeat this process for $100$ episodes. With
each iteration, the weights got updated and the model learned to
classify dynamically. We built the neural network model using keras
\cite{keras}. As keras is a Python library, we used another library
keras2cpp \cite{keras2cpp}, which allowed us to use the trained
keras model in C++ as a function.

\subsubsection{Deep Q-Learning}

In the versions $6.0$ and $6.5$, the first step to build
the sub-state was to get parameters as in equation (\ref{linreg})
for a good value of $\alpha$. This was done by first simulating
several games on different values of $\alpha$ as described in
section \ref{Q-rl}. We varied $\alpha$ from $0$ to $1$ with
increments of $1/30$ for board configuration with the larger
dimension $5$ to $10$ and $20$ with mine ratio from $0.5$ to $0.25$
playing $100$ games in each. We then ran linear regression as in
equation (\ref{linreg}) with $\alpha$ which gave the maximum win
ratio with respect to the board size configuration and mine ratio.
The regression gave an r-squared value of $0.27$ and thus was
accurate.

On obtaining $\alpha$ we get the score for each block using equation
(\ref{P-sc}) which we could then used to build sub-states of
$sub=3$. For running the simulation to obtain discounted rewards
several games were played where the net reward and the
action-sub-state pair was recorded. Each reward was
calculated as given in section \ref{Q-rl}.

For the discount, we have used a linearly increasing discount over
an exponentially increasing discount so as to punish actions which
may lead to a loss severely. This converges as all our rewards are
between $0-1$, but the discounting factor can increase to any
number.

To obtain the simulations, as in section \ref{ml}, we have used
manhattan distance as a base heuristic. We simulated various games with
$5$ to $20$ and $30$ with mine ratios from $0.05$ to $0.30$ and
playing $200$ different games on each configuration. For each game,
we recorded the sequence of a sub-state-reward pair, where the
reward was the full discounted reward calculated after the game was
over and the sub-state is a vector of length $9$ denoting the sub-state.

Since the sub-state captures information from both the state and
action, a neural network of the configuration in the table (\ref{rl})
built using keras and was trained to learn the sub-state
vector-reward mapping with mean squared error as the loss and $rmsprop$
as the weight optimizer. Using this as the first training pass, we
then used the model obtained above to get more reward-sub-state
pairs from the model obtained after training as described above and
trained the model in an iterative model. The iterative training was
done on $50$ episodes of the larger dimension of $5$ to $20$ and
$30$ with mine ratios from $0.05$ to $0.20$ and playing $5$ games on
each configuration. This gave us the iterative pass trained model.

\subsection{Results}

In this section, we study the performance of each version by
simulating them on several minesweeper boards. In this way, we can
have exact performance statistics for each version for their
interpretation and comparison. We compare the win ratios of these
methods and their dependencies on the mine ratio, the number of blocks and
the dimensional ratio of the board. The performance data is collected by
simulation of games where board size is changing from $5$ to $15$
with dimensional ratio $0.5$ (rectangular-board) to $1$
(square-board). For all these configurations of the boards, the mine
ratio is set to vary from $5\%$ to $25\%$ with time out of $5$
seconds for each game. As the winning accuracy almost converges to $0$ after decreasing the
density of mines reaches around $25\%$ in any minesweeper board, we
can get an illustrative view of this logistics behaviour.

\begin{figure}
\includegraphics[scale=0.5]{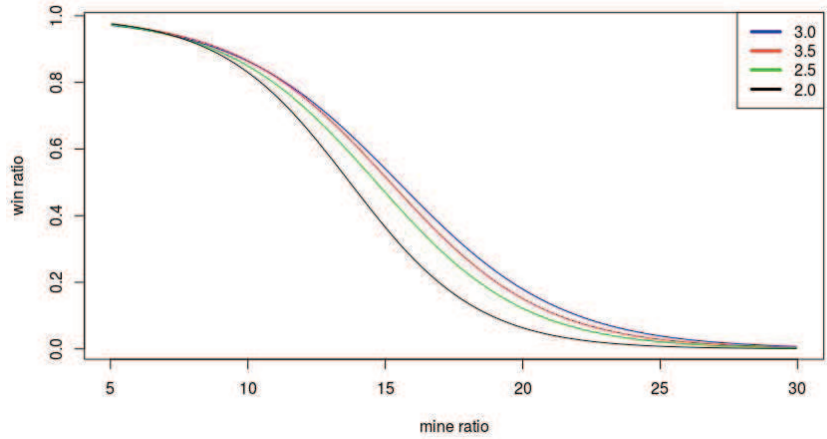}
\caption{Mine ratio vs. win ratio in CSP based
algorithms}\label{2_35}
\end{figure}

As of version $1.0$, it is simple backtracking, it takes a lot of
time to execute and that\'s why we do not consider it for comparison
with other algorithms. In fig. \ref{2_35}, we compare the winning
accuracy of version $2.0$ with versions up to $3.5$ with respect to
different mine ratios. We can interpret from the figure that,
version $2.0$ performs the worst, as it takes more than $5$ seconds (computationally very slow)
to solve that causes frequent times out. The accuracy of versions
$2.5$ and $3.5$ are very close to that of $2.0$ and $3.0$ due to
limited traversal. Overall, version $3.0$ performs much better
than others such as version 2.0, version 2.5 and version 3.5.

\begin{figure}
\includegraphics[scale=0.5]{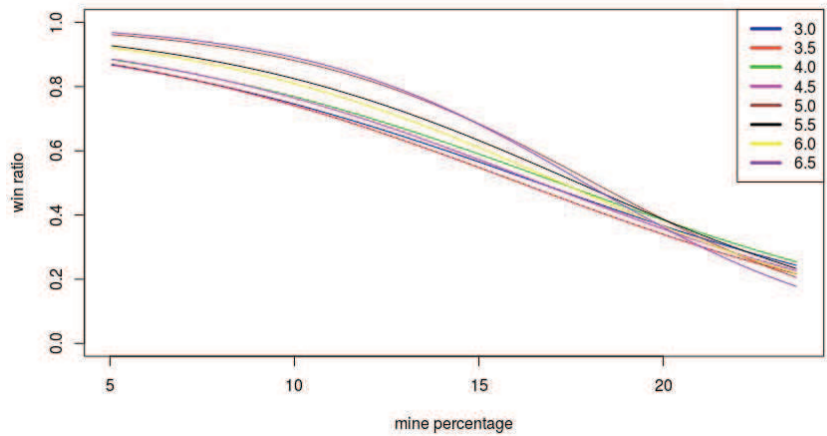}
\caption{Mine ratio vs. win ratio for all versions in table
(\ref{versions})}\label{3_65}
\end{figure}

\begin{figure}
\includegraphics[scale=0.4]{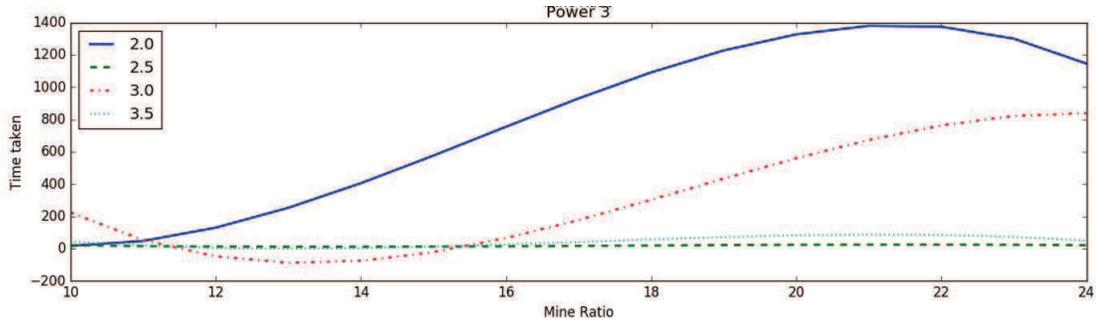}
\caption{Time vs mine percentage; Part of the graph goes to negative
time due to cubic regression limits}\label{time}
\end{figure}

This difference in time taken is also illustrated in fig. \ref{time}
which shows how much slower is version $2.0$ as compared to version
$3.0$. Notice that with very high mine percentages, version $2.0$
begins to lose fast (as the games become harder) making it less
accurate. The same figure also shows that the limited versions such
as version $2.5$ and version $3.5$ are much faster than the others,
with a very low difference in accuracy as seen from fig. \ref{2_35}.
Noted that the negative values of time are a result of cubic
regression. We compared the performance of version $3.0$ with other
versions to analyze the effects of adding heuristics in fig.
\ref{3_65} for same Minesweeper boards. The heuristic in section
\ref{manD} is experimentally proved to be useful for better results
as we notice an improvement in version $4.0$ over $3.0$ which is
also evident from fig. \ref{2_35}.

If we talk about MDP formulations, one of the heuristics we use is
the number of mines left for a state in a game that results in
further improvement in the accuracy. Now, as the number of total
mines, $n$, is increased for a board, that is, the mine ratio increase,
the effect of left mines on the result decreases. Hence, versions
from $5.0$ to $6.5$ also produce similar results as previous
versions and this was also seen in fig. \ref{3_65} with mine
ratio above $19\%$. Overall, due to over-fitting, version $5.0$ and
$6.5$ outperforms all other versions with a visible difference with
$5.5$ and $6.5$ following them.

\begin{figure}
\includegraphics[scale=0.5]{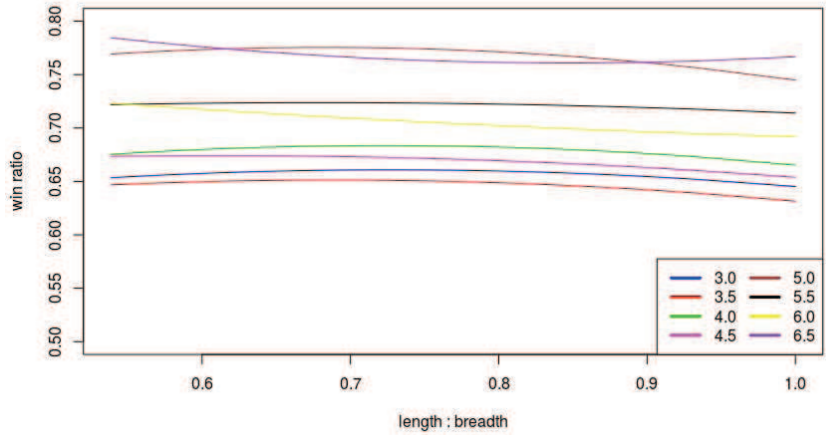}
\caption{Win ration vs ratio of length to breadth ($p:q$) of the
minesweeper board. Lower ratios indicate more ``rectangularly" as
opposed to higher ratios representing "squareness"}\label{p/q}
\end{figure}

\begin{figure}
\includegraphics[scale=0.5]{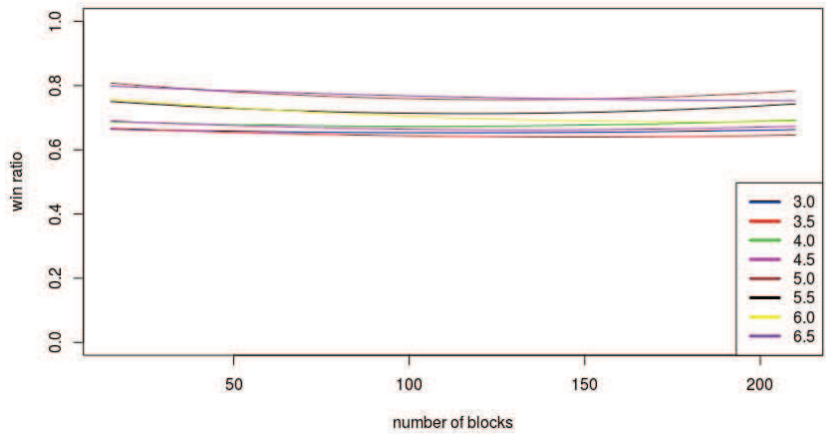}
\caption{Win ratio vs Number of total blocks in minefield. ($p \times q$) }\label{p*q}
\end{figure}

We also analyze the relationship between the win ratio of these
algorithms with a number of blocks in the board, $p\times q$ in fig.
\ref{p*q} as well as dimensional ratio, $\frac{p}{q}$ in fig.
\ref{p/q} and find that it does not affect the overall accuracy of
any versions as such.

The mine configuration of size $9 \times 9$ with $10$ mines is
considered as a standard beginner size board. The accuracy obtained
by algorithms presented in this paper compared to \cite{ucthcsp} is
given in Table \ref{compare_all}.
\\
\\
\begin{table}[h!]
      \centering
      \caption{Standard Beginner Board Accuracy} \label{compare_all}
      \begin{tabular}{|c|c|}
       \hline
      \multicolumn{ 1}{|c|}{{\bf Version Number}} & {\bf Accuracy} \\
      \hline
         $2.0$ &  $82.14$ \\
      \hline
         $2.5$ &  $77.34$ \\
      \hline
         $3.0$ &  $82.20$ \\
      \hline
         $3.5$ &  $82.20$ \\
      \hline
         $4.0$ &  $86.24$ \\
      \hline
         $4.5$ &  $86.22$ \\
      \hline
         $5.0$ &  $92.02$ (max)\\
      \hline
         $5.5$ &  $89.21$ \\
      \hline
         $6.0$ &  $87.58$ \\
      \hline
         $6.5$ &  $91.41$ \\
      \hline
         OH \cite{ucthcsp} &  $89.9$ \\
      \hline
      \end{tabular}
\end{table}
\\
\section{Conclusion}\label{Conclusion}

In this paper, we have used the realization of the Minesweeper game
as an MDP and as a CSP to create several kinds of solving methods.
We introduced DSScsp (Algorithm \ref{DSScsp}), which can enumerate
all solutions of the CSP much faster than backtracking while being
just as accurate. We also introduce the concept of limited traversal
which obtains a subset of all possible solutions and showed that it
gives nearly as accurate results as full traversal, while being much
faster. We improved preexisting heuristics (Version $4.0$) and
introduced new ones which used machine learning and deep Q-learning
(Version $5.0$ \& version $6.5$) on the MDP formulation of the game
which we showed to be better than that of existing methods (Table
(\ref{compare_all})). We also showed that deep Q-learning is
marginally better than machine learning (classification) methods in
general, but not always as can be seen in the Table
(\ref{compare_all}). Overall, we claim that our method of deep
Q-learning (Version $6.5$) can play Minesweeper to a high degree of
accuracy while still being fast enough to play boards as large as
$200$ total blocks.

\vskip 0.2in
 \bibliography{CSP}
 \bibliographystyle{plain}

\end{document}